\UseRawInputEncoding
\documentclass[letterpaper, 10 pt, conference]{ieeeconf}  

\IEEEoverridecommandlockouts                              

\usepackage{graphics} 
\usepackage{epsfig} 
\usepackage{mathptmx} 
\usepackage{times} 
\usepackage{amsmath} 
\usepackage{amssymb}  
\usepackage{booktabs}
\usepackage{marvosym}

\makeatletter
\let\NAT@parse\undefined
\makeatother
\usepackage[colorlinks]{hyperref}

\title{\LARGE \bf
Learning Structural Latent Points for Efficient Visual Representations in Robotic Manipulation
}

\author{Yicheng Jiang\textsuperscript{1,*}, Jiaxu Wang\textsuperscript{1,3,*}, Junhao He\textsuperscript{2}, Zesen Gan\textsuperscript{1}, Junhao Li\textsuperscript{1}, \\
Qiang Zhang\textsuperscript{2,4}, Jingkai Sun\textsuperscript{5}, Jiahang Cao\textsuperscript{5}, Mingyuan Sun\textsuperscript{6}, Xiangyu Yue\textsuperscript{3,\Letter} and Qiming Shao\textsuperscript{1,\Letter}
\thanks{\textsuperscript{*} Contributed equally co-first authors, order determined by coin toss.}
\thanks{\textsuperscript{\Letter}Corresponding authors.}
\thanks{\textsuperscript{1}The Hong Kong University of Science and Technology}%
\thanks{\textsuperscript{2}The Hong Kong University of Science and Technology (Guangzhou)}%
\thanks{\textsuperscript{3}MMLab, The Chinese University of Hong Kong}%
\thanks{\textsuperscript{4}X-Humanoid Robots}%
\thanks{\textsuperscript{5}The University of Hong Kong}
\thanks{\textsuperscript{6}Tsinghua University}
}

\begin{document}

\maketitle
\thispagestyle{empty}
\pagestyle{empty}

\begin{abstract}
Current 3D-aware pretraining methods for embodied perception and manipulation are largely built on differentiable rendering frameworks, producing either fully implicit neural fields or fully explicit geometric primitives. Implicit representations, while expressive, lack explicit structural cues, whereas explicit ones preserve geometry but suffer from resolution limits and weak generalization. To address these limitations, we propose a novel pretraining framework that learns a hybrid representation-structural latent points. Specifically, we insert a point-wise latent variational autoencoder into the latent space of a point-cloud autoencoder, jointly regularizing point-wise features and coordinates toward a Gaussian prior. The resulting compact latent preserves coarse structural tendencies, which do not encode precise geometry but capture richer rough shape and semantic information, effectively combining the expressiveness of implicit representations with the structural priors of explicit ones. In addition, informed by shared design choices in prior work, we develop a streamlined, efficient 3DGS-based rendering pipeline that is deliberately kept lightweight, improving efficiency while leaving greater representational capacity to the front-end latent module. Extensive evaluations on RLBench, ManiSkill2, and a real-robot platform demonstrate consistent gains in task success, sample efficiency, and robustness to viewpoint and scene variations over strong baselines. Ablation studies further confirm that each component of our framework is critical to overall performance.
\end{abstract}

\section{Introduction}


Early approaches to embodied intelligence fed raw sensory observations directly into policy networks. Although simple, this strategy was soon found to be inefficient and brittle. A more effective paradigm emerged by introducing general-purpose feature extractors that compress raw inputs into compact latent representations, improving both efficiency and downstream task performance. However, such models primarily capture 2D features and thus fall short in supporting agents that must act in the 3D physical world.

To overcome this limitation, recent works have developed various 3D-aware pretraining frameworks that exploit multi-view observations. By incorporating 3D rendering mechanisms, such as Neural Radiance Field (NeRF) \cite{mildenhall2021nerf} or 3D Gaussian Splatting (3DGS) \cite{kerbl20233d} , into the training process, these methods encourage models to acquire geometry-aware representations. NeRF-based approaches typically aggregate multi-view features into a volumetric representation, which is then rendered via ray casting for self-supervised training. The resulting representation is entirely implicit, encoding spatial and semantic information in a continuous volumetric field, but without exposing any structured or explicit geometric information. While such implicit embeddings are expressive, they lack interpretable structural cues that can be directly exploited by downstream embodied agents. Moreover, NeRF-based pretraining is computationally expensive due to the inefficiency of ray casting.

In contrast, other studies leverage the high efficiency of 3DGS to build the pretraining pipeline. These methods generally predict the properties of Gaussian splats from multi-view images and adopt the resulting splats directly as explicit scene representations, sometimes further enriching each splat point with semantic features. Compared with implicit fields, such explicit point-like structures naturally encode interpretable geometric information, which can benefit downstream reasoning and manipulation tasks. However, their capacity is constrained by resolution: to maintain lightweight policy networks, point-like splats are often downsampled, leading to the loss of fine-grained structural information.
To conclude, implicit methods offer strong expressiveness, while explicit methods provide well-defined structural priors. 

Another challenge with the aforementioned training paradigms is that their performance often degrades when transferring to downstream tasks whose scenes differ significantly from the pretraining domain. During pretraining, the projected point clouds tend to occupy a well-structured and dense latent space, leading to strong representation quality. However, when downstream scenes deviate too much, their projections can lie far from the pretrained latent manifold, resulting in poor representations and weaker task performance.

To combine the strengths of implicit representations in
expressiveness and explicit representations in structural information, we propose a framework that learns structural
latent points. 
Specifically, our method integrates a standard point autoencoder with a point-wise latent variational autoencoder (PL-VAE). The PL-VAE operates on the sparse feature point cloud residing in the latent space of a U-Net-like point cloud autoencoder, and regularizes both the spatial distribution of points and their associated features toward a Gaussian prior. Through this process, the sparse feature points are transformed into a latent representation that remains structured yet no longer strictly explicit. Rather than encoding precise geometry, it captures structural information in a probabilistically regularized form, thereby serving as a structural implicit representation.
This design preserves geometric
tendencies while retaining the compactness and flexibility of
latent representations, thus bringing together the best of both
paradigms.


We further adopt 3DGS to construct our self-supervised pretraining pipeline due to its computational efficiency. Unlike prior frameworks that often introduce complex and redundant architectural designs, we deliberately remove unnecessary components and retain only the core modules that are consistently shared across various approaches and proven to be effective. Our goal is to design a streamlined and lightweight pretraining pipeline that is both simple and effective. 
We extensively evaluate this method across RLBench and the ManiSkill2 platform and compare it with several baselines. The results show that our method achieves the best overall performance. We further validate the practicality of our pretrained framework on a real-world robotic platform across six diverse experimental tasks. Moreover, we conduct ablation studies for all necessary designs. This confirms the effectiveness of our structural latent point representation in enhancing both generalization and performance in downstream embodied tasks.

\section{Related Work}
\subsection{Visual Representation Learning}
Recent advances in visual representation learning have greatly influenced robot learning by enabling robots to better perceive and interact with their environments. However, developing robust visual representations for robotics remains both critical and challenging. Some methods \cite{bharadhwaj2024roboagent,brohan2022rt, jang2022bc, radosavovic2023robot} leverage task-specific data, such as real-world robot demonstrations, but the high cost of data collection limits scalability. Other approaches \cite{bharadhwaj2024roboagent,yarats2021image,laskin2020reinforcement,yu2023scaling} turn to data augmentation, self-supervised learning, and the incorporation of task priors to reduce this dependency. Meanwhile, large-scale pretraining on unlabeled data, like \cite{he2022masked,radford2021learning, oquab2023dinov2, laskin2020curl,huang2023ponder}, has substantially improved transfer learning and enabled zero-shot capabilities. However, these approaches focus primarily on 2D representations, which may be suboptimal for robots operating in 3D spaces. To address this gap, we propose a framework for learning 3D visual representations: by learning geometric and semantic properties through latent point clouds, our method enhances robotic perception in 3D environments and improves performance in various downstream tasks.

\subsection{3D-aware Pretraining for robotics and embodied AI}

The development of visual pretraining \cite{bharadhwaj2024roboagent,laskin2020reinforcement,yu2023scaling,radford2021learning,he2022masked,oquab2023dinov2} has shown an impressive generalization ability on robotic tasks.  However, 2D pretraining approaches still suffer from the lack of spatial information. To address this issue, some works \cite{wang2024lift3d, ke20243d, goyal2024rvt, niu2025pre,gervet2023act3d} typically attempt to lift 2D priors into 3D spaces and introduce spatial awareness into 2D foundation models. SPA \cite{zhu2024spa} leverages neural volume rendering on multi-view images to endow a vanilla ViT with intrinsic 3D spatial awareness, but requires huge computing resources and training time to achieve the best performance. Lift3D \cite{jia2024lift3d} leverages 2D foundation models to guide the representation learning, but it still relies on the completeness of the point cloud observation.


\subsection{Neural Rendering}



Neural volume rendering-pioneered by NeRF \cite{mildenhall2021nerf}-provides a fully differentiable bridge between 2D views and 3D scenes, and has since been accelerated via proposal networks and factorization or by swapping in alternative implicit surfaces \cite{ortiz2022isdf,wang2022go, huang2023ponder, wang2024learning, wang2024evggs}. However, NeRF-based volumetric rendering methods often require sampling a large number of points along each ray, which incurs high computational cost and leads to slow convergence during training. More recently, 3D Gaussian Splatting (3DGS) \cite{kerbl20233d} has emerged as an explicit, high-fidelity scene representation that splats learned Gaussian kernels into a volumetric field for ultra-fast rendering and incremental updates, and Feature Gaussian Splatting \cite{qiu2024feature,zhou2024feature,qin2024langsplat} further extends this by encoding rich point-wise features into each Gaussian-enabling not only real-time view synthesis but also efficient multi-view feature aggregation for downstream tasks.

\begin{figure*}[t]
    \centering    
    \includegraphics[width=0.92\linewidth]{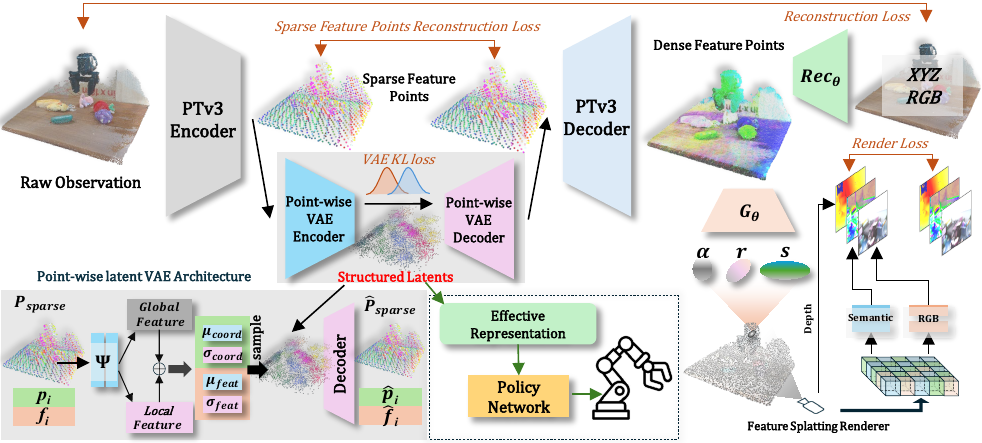}
    \vspace{-3mm}
    \caption{The overview of our main pipeline.}
    \label{fig: main pipeline}
    \vspace{-5mm}
\end{figure*}

\section{Preliminaries}
\subsection{3D-aware Visual Representation Learning}
Visual pretraining has traditionally focused on learning representations from 2D images, where encoders are trained to capture semantic and appearance information useful for downstream tasks. While effective in purely visual domains, such representations often lack awareness of the 3D structure that is essential for embodied agents operating in physical environments. To inject 3D awareness into the learned features, recent works typically resort to differentiable 3D rendering pipelines, which enable self-supervised training by reconstructing views from multi-view images and camera poses. These pipelines provide the supervisory signal needed for encoders to acquire geometry-aware representations from purely 2D observations.

A representative paradigm is the NeRF-style framework. Let $\{I_i\}_{i=1}^N$ denote multi-view images and camera parameters $\{T_i\}_{i=1}^N$. A unified abstraction for NeRF-style pretraining can be expressed as
\begin{equation}
    \hat{I}_k = \Psi_d\!\Big({R}_{\mathrm{ray}}\big(\Psi_e(I_{i=1}^N), T_k\big)\Big),
    \label{eq: nerf style}
\end{equation}
where $\Psi_e$ is a multi-view feature extractor (e.g., a 3D CNN) that constructs a 3D feature volume, and $R_{\mathrm{ray}}$ denotes the NeRF renderer. The queried features are aggregated and decoded into the image domain by $\Psi_d$, and the synthesized image $\hat{I}_k$ is supervised against the ground truth $I_k$. The latent volume $\Psi_e(I_{i=1}^N)$ then serves as the scene representation, which is compact and expressive.  However, this representation remains entirely implicit, lacking explicit structural information that could be directly connected with downstream tasks. In addition, the ray-casting operation is computationally demanding.

Another paradigm builds upon the 3DGS pipeline. Feature maps are first extracted from each image and then projected into 3D space to form a feature point cloud. A point-based network predicts Gaussian splat parameters, and the scene is rendered via the GS renderer:
\begin{equation}
\hat{I}_k = R_{\mathrm{rast}}\Big(\Phi_p\big(\pi_{k}\ast(\Phi_e(I_{i=1}^N), T_k)\big)\Big),
\label{eq: gs style}
\end{equation}
where $\Phi_e$ is a pretrained image encoder, $\pi_k$ is the unprojection function under camera $T_k$, and $\Phi_p$ predicts the splat properties. Prior work has used either the predicted Gaussian parameters or the intermediate feature point set as the scene representation. $R_{\mathrm{rast}}$ is highly efficient. Nevertheless, the explicit and point-based nature inherently constrains them by resolution and limits their ability to capture compact, high-level abstractions of scene geometry.

Our framework \ref{sec. main architecture} incorporates a point-wise latent VAE into the latent space of a standard Point Transformer v3 (PTv3) to learn a structural latent representation, which is trained with a simple yet effective 3DGS-based self-supervised paradigm to endow the model with 3D awareness. The resulting representation combines the expressiveness of implicit methods with the structural cues of explicit ones, producing a compact, geometry-aware abstraction that is particularly well suited for downstream embodied tasks.

\subsection{Gaussian Splatting}
\label{sec. 3dgs}
3DGS represents a 3D scene using a set of Gaussian primitives, each defined by a centroid $u \in \mathbb{R}^3$, and a set of properties related to its local shape, i.e. $\mathbf{G}_i=\{\Sigma_i=\mathbf{R}_{g,i}\mathbf{S}_{g,i}\mathbf{S}_{g,i}^{T}\mathbf{R}_{g,i}^{T}, o_i, c_i\}$
in which, to ensure stable backpropagation during optimization. The lower subscript $i$ refers to the $i-th$ element in the Gaussian splat set. $\Sigma_i$ is decomposed into a rotation matrix $\mathbf{R}_{g,i}$ and scale matrix $\mathbf{S}_{g,i}$. $o_i,c_i$ denote the opacity and color of the $i-th$ splat. To preserve differentiability in the rendering process, a local affine approximation is applied by linearizing the projection using the Jacobian matrix $J$, i.e. $\Sigma^{'}=JW\Sigma W^{T}J^{T}$
Then, the final color and depth map is defined via:
\begin{equation}
    \mathbf{C} = \sum_{i\in \mathcal{N}}c_i\alpha_i\prod_{j=1}^{i-1}(1-\alpha_j), ~~
    \mathbf{D} = \sum_{i\in \mathcal{N}}d_i\alpha_i\prod_{j=1}^{i-1}(1-\alpha_j),
    \label{eq: rendering}
\end{equation}
where$d_i$ refers to the distance between $i-th$ splat and camera center. $\alpha_i$ denotes the soft occupation of the splat at the corresponding 2D location of the image plane, which can be obtained by $\alpha_i(x) = o_iexp(-\frac{1}{2}(x-\mu_i)^T{\Sigma}_{i}^{T}(x-\mu_i))$.

\section{Methodology}
\subsection{Main Architecture of the Pretraining Framework}
\label{sec. main architecture}

The overall architecture of our pretraining framework can be divided into two main components. The first component focuses on point cloud encoding and decoding for effective structured latents extraction, while the second component performs geometric reasoning and 2D feature rendering for multi-view supervision.

In the first stage, we adopt Point Transformer V3 (PTv3) together with a point-wise latent VAE. The raw point cloud $\mathbf{P}_{\text{raw}}$ is first encoded by the PTv3 encoder to obtain the sparse feature points, marked as: $
\mathbf{P}_{\text{sparse}} = \text{PTv3}_{\text{enc}}(\mathbf{P}_{\text{raw}})$.

Typically, the sparse feature point cloud is directly fed into a PTv3 decoder that upsamples it to recover the original resolution, producing a dense feature point cloud. In contrast, the key distinction of our approach is that, before decoding, we insert a \textbf{point-wise latent VAE} (\textbf{PL-VAE}) into the encoder’s latent space. This additional reconstruction step is designed to enforce a compact and robust representation, ensuring stronger generalization for downstream tasks, which we will introduce in the next section in detail. The output of the \textbf{PL-VAE} is referred to as $\hat{\mathbf{P}}_{\text{sparse}}$ and passed to the decoder, which restores the point cloud resolution, stated as 
$
\mathbf{P}_{\text{dense}} = \text{PTv3}_{\text{dec}}(\hat{\mathbf{P}}_{\text{sparse}}).
$

The second component aims to equip the framework with 3D awareness through differentiable rendering. 

This rendering component is deliberately kept simple, it includes only the common, effective, and lightweight modules inspired by prior 3DGS-based representation learning \cite{wang2024pfgs, wang2024gsrl, qiu2024feature}. Keeping the renderer simple ensures the first component remains the primary contributor of representation learning, preventing the second component from absorbing excessive burden or compensating for the encoder. In turn, this encourages the structural latent module to learn more informative and transferable features.

Concretely, the component comprises a geometric reasoning module \(G_\theta\), a feature-splatting module, and a reconstruction module \(Rec_\theta\). The module \(G_\theta\) predicts point-wise Gaussian properties, as described in Sec.~\ref{sec. 3dgs}, from the decoded dense feature points:
\begin{equation}
\mathbf{G} \;=\; G_{\theta}\!\left(\mathbf{P}_{\text{dense}}\right).
\end{equation}
Reasoning over these spatial attributes enables 3DGS rendering, effectively bridging the PTv3 latent space with local geometry \cite{wang2024pfgs, zhou2025gps, wang2024evggs}. We then perform \emph{feature splatting}, which extends conventional Gaussian rasterization to render point-wise feature information (and depth) rather than color.
Given camera v, we splat features and depths only, stated in Eq.~\ref{eq: feature splat}, rather than both features and color. 
\begin{equation}
\bigl(\hat{\mathbf{F}}_{v},\, \hat{\mathbf{D}}_{v}\bigr) \;=\; R_{\mathrm{v}}\!\left(\mathbf{G},\, \mathbf{P}_{\text{dense}}\right).
\label{eq: feature splat}
\end{equation}
Prior works vary in this choice, but ablations consistently show that representation quality hinges on feature splatting, while color splatting mainly improves image fidelity \cite{zhou2024feature, qiu2024feature}. Since our goal is to learn geometry-aware features (not photorealistic images), feature-only splatting keeps the pipeline concise and focuses model capacity on representation learning.
The splatted feature maps are then decoded by two projectors into semantic features and color:
\begin{equation}
\hat{\mathbf{I}}_v,\hat{\mathbf{F}}_v^s=H_{c}(\hat{\mathbf{F}_v}), H_{s}(\hat{\mathbf{F}_v}).
    \label{eq: post feature decoding}
\end{equation}
Unlike methods that render feature and color separately and align them post hoc, we apply both supervisions directly to a shared latent feature space, which simplifies the pipeline and tightly couples semantic and appearance cues, yielding stronger representations. We also align rendered and ground-truth depths, providing direct gradients to \(G_\theta\).

In addition to rendering alignment, we introduce a lightweight reconstruction module that regresses original color and 3D coordinates of each point from the decoded dense features. We add this module because, early in training, the rendered features are underdeveloped and provide relatively indirect supervision. The auxiliary reconstruction loss anchors the representation to basic appearance and geometry, accelerating convergence, stabilizing the initial training phase, and steering the features toward a well-behaved latent space for more robust learning overall.

\subsection{Point-wise Latent VAE}
\label{sec. Point-wise VAE}
Unlike traditional Variational Auto Encoder (VAE), which only produces a global feature for the point cloud, the PL-VAE operates both at the feature and coordinates of the latent sparse feature points ($\mathbf{P}_{sparse}$), constructing a 3D-aware structured latent. 

For the encoder of the PL-VAE, we extract global and local descriptors from these sparse points: 
\begin{equation}            
 f_{global}, f_i =\mathbf{Agg}(\Psi(\mathbf{P}_{sparse})),\mathbf{MLP}(\Psi(\mathbf{P}_{sparse})),
    \label{eq: plvae1}
\end{equation}
where $\mathbf{Agg}$ denotes the aggregation based on attention pooling. $\Psi$ is a PointNet backbone for extracting features. The global features provide a holistic overview of the scene, while local features capture point-wise context. 

For each point, we define a joint representation $h_i=cat(f_{global},f_{i})$ by concatenating the $f_{\text{global}}$ and $f_i$. This joint representation $h_i$ is then used to predict the Gaussian distribution parameters in the VAE, covering both the feature dimensions and the coordinate dimensions for each point.
\begin{equation}
    (\mu^f_i, \sigma^f_i),(\mu^p_i, \sigma^p_i)=\phi_f(h_i),\phi_p(h_i),
    \label{eq: regress vae gaussian parameters}
\end{equation}
where $\phi_f(\cdot)$ and $\phi_p(\cdot)$ are MLPs that output the mean and standard deviation for the respective distributions.
We then sample latent variables using the reparameterization trick to form the intermediate features, that are regularized to lie close to a Gaussian distribution in the VAE latent space, referred to as:
$
    {\mathbf{z}_{vae}=(z^f_i, z^p_i)}_{i=1}^N.$
Compared with the $\mathbf{P}_{sparse}$, the $\mathbf{z}_{vae}$ are more compact and closer to the Gaussian prior. The point positions no longer represent precise geometry; instead, they behave like Gaussian-smoothed approximations. Such approximate structural encoding is advantageous in the latent space: it reduces sensitivity to noise and fine-grained variations while preserving the overall geometric tendencies, leading to smoother manifolds and stronger generalization for downstream tasks.
Then, $\mathbf{z}_{vae}$ is decoded via a VAE decoder to reconstruct $\hat{\mathbf{P}}_{sparse}$, enabling cascaded connections within the overall framework and facilitating end-to-end joint optimization:
$
\hat{\mathbf{P}}_{sparse}= \Psi_{\text{dec}}(\mathbf{z}_{vae}).
$
After that, the reconstructed $\hat{\mathbf{P}}_{sparse}$ is fed to the PTv3 decoder.



\subsection{Effective Representation for Downstream Task}
\label{Sec. representation detail}
The reparameterized latent representation ${\textbf{z}}_{vae}$, obtained from the PL-VAE as described in Section \ref{sec. Point-wise VAE}, serves as the effective input for downstream embodied tasks. It provides a compact and smooth latent space regularized toward a Gaussian prior, capturing high-level structural regularities. This abstraction enhances robustness to noise and scene variations, improves generalization across domains.

In downstream applications, ${\textbf{z}}_{vae}$ can be directly used as the input to a policy network, bypassing the need for explicit geometric reconstruction. For example, in robot imitation learning, the policy network $\pi$ receives the latent point cloud representation ${\textbf{z}}_{vae}$ along with the robot's proprioceptive state ${O}_{agent}$ (such as joint positions, velocities, or gripper status), and outputs a corresponding action $\hat{a}$:

\begin{equation}
    \hat{a}= \pi(\mathbf{z}_{vae} , {O}_{agent}).
    \label{eq. zvae downstream}
\end{equation}


\subsection{Optimization Strategies}
\label{Sec. loss}
This section describes the optimization strategies and loss functions employed in this framework.

According to Sec.~\ref{sec. main architecture} and \ref{sec. Point-wise VAE}, the fundamental loss is the rendering alignment of the decoded RGB and semantic feature map, as well as the rendered depth map (Eq.~\ref{eq: feature splat} and \ref{eq: post feature decoding}). This can be described as:
\begin{equation}
    \begin{aligned}
        L_{render} = \frac{1}{V} \sum_{i=1}^{V} \Big( 
            & \beta_1 \cdot \| \mathbf{I}_v - \hat{\mathbf{I}}_v \|_1
          +\, \beta_2 \cdot \| \mathbf{D}_v - \hat{\mathbf{D}}_v \|_1 \\
          & +\, \beta_3 \cdot \| \mathbf{F}_v^s -  \hat{\mathbf{F}}_v^s \|_1
        \Big).
\end{aligned}
\label{eq:rendering_loss}
\end{equation}
The symbols here retain the same meaning as the above text. The groundtruth of semantic maps $\mathbf{F}_s^v$ is produced by E-RADIO \cite{ranzinger2024amradio}. In our experiments, we set $\beta_1=1,\beta_2=0.2,\beta_3=0.1$ constantly.

Moreover, we introduce an auxiliary reconstruction loss to map the $\mathbf{P}_{dense}$ to the original $\mathbf{P}_{raw}$, thereby improving both convergence speed and training stability.
\begin{equation}
L_{recon}=\frac{1}{N}\sum_{i=1}^{n}(\|p_i-\hat{p_i} \|_1 + \|c_i-\hat{c_i} \|_1).
    \label{eq. recon loss}
\end{equation}
The $N$ denotes the number of points, $p_i,c_i$ represent the raw point coordinate and color. 


Additionally, we introduce two loss terms to supervise the PL-VAE depicted. The first term is the L1 latent reconstruction loss, aiming to reconstruct the $\mathbf{P}_{sparse}$ with VAE:
\begin{equation}
    L_{vae}=\frac{1}{M}\sum_m^1(\omega_1\|p_i-\hat{p_i}\|+\omega_2\|f_i-\hat{f_i}\|),
\label{eq:vae}
\end{equation}
in which $p_i,f_i$ are the coordinates and features of the $\mathbf{P}_{sparse}$. We use $\omega_1$ and $\omega_2$ to balance the reconstruction of point positions and features, since, as shown in Eq.~\ref{eq: regress vae gaussian parameters}, they are sampled from Gaussian parameters predicted by different heads and thus exhibit different statistical scales. In our experiments, the $\omega_1=1,\omega_2=0.1$ Conventional VAEs for point generation often employ the Chamfer Distance loss to align the overall distribution of points. However, their objective is typically generation rather than exact recovery. In our case, the goal is to precisely reconstruct $\mathbf{P}_{sparse}$, and therefore we adopt an L1 loss instead.

Furthermore, the classic VAE KL loss is introduced as well:
\begin{equation}
    \begin{aligned}
        L_{KL} = \text{KL}(\mathcal{N}(\mu^f_i, \sigma^f_i) \,\|\, \mathcal{N}) 
          +\, \text{KL}\big(\mathcal{N}(\mu^p_i, \sigma^p_i) \,\|\, \mathcal{N})
        ).
    \end{aligned}
\label{eq: kl loss}
\end{equation}
This term encourages both the feature- and coordinate-level posteriors to remain close to a standard Gaussian prior $\mathcal{N}$, promoting a smooth and well-behaved latent space. Therefore, the total loss terms can be described as:
\begin{equation}
    L_{total}= (1-w(t))L_{render}+w(t)L_{recon}+L_{vae}+L_{KL}.
    \label{eq: total loss}
\end{equation}
Here $w(t)$ is an annealing weight that starts at $1$ and gradually decays to $0.1$. This scheduling emphasizes reconstruction early, when rendered features are immature, and progressively shifts focus to rendering alignment as training stabilizes. 

\section{Experiments}
\noindent \textbf{Benchmarks}. In this section, we present overall evaluations of the proposed pretraining methods. During recent years, the development of robotic simulators has significantly enhanced for reproducible and efficient robot performance evaluation. Hence, we selected two widely-adopted simulator benchmarks: RLBench \cite{james2020rlbench}, and Maniskill2 \cite{gu2023maniskill2}. The two benchmarks are chosen to be diverse and representative. Furthermore, we establish a real-world robot platform to validate the effectiveness of our method on real-world cases. Our goal is to evaluate the performance of the pretrained representation; therefore, we uniformly pretrain all baselines on the same dataset ScannetV2 before we adopt them for the downstream manipulation tasks.

\noindent \textbf{Metrics}. We use the Success Rate (S.R.) to assess the performance of each method on individual tasks. Additionally, following the evaluation protocol from VC-1 \cite{majumdar2023we} and PCM \cite{zhu2024point}, we report two aggregate metrics: Mean S.R. and Mean Rank. Mean S.R. reflects the average success rate across all tasks. While mean Rank ranks methods by their success rate on each task and then averages these ranks across all tasks.

\subsection{Evaluation on Simulation benchmarks}
We evaluate our approach on RLBench and ManiSkill2 using OBSBench~\cite{zhu2024point}, which provides standardized pipelines and a diverse set of baselines within a unified framework for benchmarking robotic performance. Following the OBSBench protocol, we adopt its default settings and train an Action Chunking Transformer (ACT) policy~\cite{zhao2023learning}. For comparison, we include three backbone models: MultiViT \cite{yarats2021image}, PTv3 \cite{wu2024point}, and SPUNet \cite{contributors2022spconv} trained from scratch under an autoencoder paradigm, as well as several pretrained 3D-aware approaches, including VC-1 \cite{majumdar2023we}, MultiMAE \cite{bachmann2022multimae}, PonderV2 \cite{zhu2023ponderv2}, GSRL \cite{wang2024gsrl,wang2024query}, SPA \cite{zhu2024spa}, and Lift3D \cite{jia2024lift3d}.

\noindent \textbf{RLBench}. Table~\ref{tab:rlbench_recomputed} reports S.R. across nine RLBench tasks. We observe that 3D-aware pretrained methods consistently outperform backbone models trained from scratch, highlighting the importance of visual representation learning. Furthermore, point-based pretraining paradigms generally yield stronger results than multi-view image pipelines. Our method achieves the best overall performance, which we attribute to its integration of implicit latents with explicit structural cues.
\begin{table*}[ht]
\centering
\caption{Results on RLBench tasks under ACT policy}
\label{tab:rlbench_recomputed}
\resizebox{0.9\textwidth}{!}{
\begin{tabular}{lcccccccccc}
 \hline
Task & MultiViT & PTV3 & SpUNet & VC-1 & MultiMAE & PonderV2 & GSRL & Lift3D & No.VAE & Ours \\
 \hline
OpenDrawer    & 0.08 & 0.40 & 0.44 & 0.24 & 0.36 & 0.60 & 0.52 & 0.54 & 0.56 & \textbf{0.62} \\
SweepTo       & \textbf{1.00} & 0.96 & 0.90 & 0.96 & \textbf{1.00} & 0.96 & 0.92 & \textbf{1.00} & \textbf{1.00} & \textbf{1.00} \\
MeatoffGrill  & 0.36 & 0.68 & 0.72 & 0.12 & 0.04 & 0.72 & 0.60 & 0.74 & 0.76 & \textbf{0.80} \\
TurnTap       & 0.00 & 0.04 & 0.00 & 0.04 & 0.08 & 0.00 & 0.02 & 0.04 & 0.08 & \textbf{0.12} \\
ReachandDrag  & 0.04 & 0.04 & 0.20 & 0.20 & 0.16 & \textbf{0.28} & 0.26 & 0.16 & 0.24 & \textbf{0.28} \\
PutMoney      & 0.28 & 0.44 & 0.60 & 0.72 & 0.56 & 0.64 & 0.62 & 0.58 & 0.48 & \textbf{0.76} \\
PushButtons   & 0.14 & 0.64 & 0.00 & 0.44 & 0.48 & 0.16 & 0.40 & 0.72 & 0.68 & \textbf{0.74} \\
CloseJar      & 0.00 & 0.44 & 0.04 & 0.08 & 0.00 & 0.28 & 0.04 & \textbf{0.54} & 0.48 & 0.52 \\
PlaceWine     & 0.00 & 0.04 & 0.00 & 0.00 & 0.08 & 0.16 & 0.08 & \textbf{0.20} & 0.16 & 0.18 \\
 \hline
Mean S.R.\,$\uparrow$   & 0.21 & 0.41 & 0.32 & 0.31 & 0.31 & 0.42 & 0.38 & 0.50 & 0.49 & \textbf{0.56} \\
Mean Rank\,$\downarrow$ & 8.56 & 6.50 & 7.39 & 6.50 & 6.44 & 4.83 & 6.11 & 3.61 & 3.56 & \textbf{1.50} \\
 \hline
\end{tabular}}
\end{table*}

\noindent \textbf{Maniskill2}. For the Maniskill2 platform, we selected 3 rigid body tasks (PickCube, StackCube, TurnFaucet) and 3 soft body tasks (Hang, Fill, and Pour) for evaluation. The full results can be seen in Table~\ref{table: maniskill results}.
Across all six ManiSkill2 tasks, our method consistently delivers the most substantial and stable performance over existing pretrained baselines. This underscores the superior generalizability of our pretraining approach.

\begin{table*}[ht]
\centering
\caption{Results on ManiSkill2 tasks under ACT policy}
\label{tab:maniskill2_reorg_swapped}
\resizebox{0.9\textwidth}{!}{
\begin{tabular}{lccccccccccc}
 \hline
Task       & MultiViT & PTV3 & SpUNet & VC-1 & MultiMAE & SPA & PonderV2 & GSRL & Lift3D & No.VAE & Ours \\
 \hline
PickCube   & 0.04 & 0.76 & 0.74 & 0.77 & 0.52 & 0.62 & 0.87 & 0.76 & 0.90 & 0.82 & \textbf{0.92} \\
StackCube  & 0.00 & 0.08 & 0.22 & 0.06 & 0.30 & 0.00 & 0.35 & 0.28 & 0.35 & 0.32 & \textbf{0.38} \\
TurnFaucet & 0.35 & 0.00 & 0.39 & 0.42 & 0.37 & 0.45 & 0.30 & 0.28 & 0.45 & 0.44 & \textbf{0.48} \\
Hang       & 0.84 & 0.88 & 0.84 & 0.84 & 0.77 & 0.92 & 0.83 & 0.88 & \textbf{1.00} & 0.96 & 0.98 \\
Pour       & 0.00 & 0.08 & 0.10 & 0.04 & 0.00 & 0.00 & 0.11 & 0.12 & 0.15 & 0.10 & \textbf{0.20} \\
Fill       & 0.76 & 0.56 & 0.66 & 0.78 & 0.68 & 0.86 & 0.73 & 0.70 & \textbf{0.90} & 0.88 & 0.89 \\
 \hline
Mean S.R.\,$\uparrow$   & 0.33 & 0.39 & 0.49 & 0.49 & 0.44 & 0.48 & 0.53 & 0.50 & 0.63 & 0.59 & \textbf{0.64} \\
Mean Rank\,$\downarrow$ & 8.92 & 8.17 & 7.42 & 6.67 & 8.67 & 6.67 & 5.92 & 6.50 & 1.83 & 3.92 & \textbf{1.33} \\
 \hline
\end{tabular}}
\label{table: maniskill results}
\vspace{-5mm}
\end{table*}

\subsection{Further Analysis and Ablation Studies}
In this section, we explain a more in-depth analysis of the proposed method and ablate all necessary components. 

\begin{figure}
    \centering
    \includegraphics[width=0.5\textwidth]{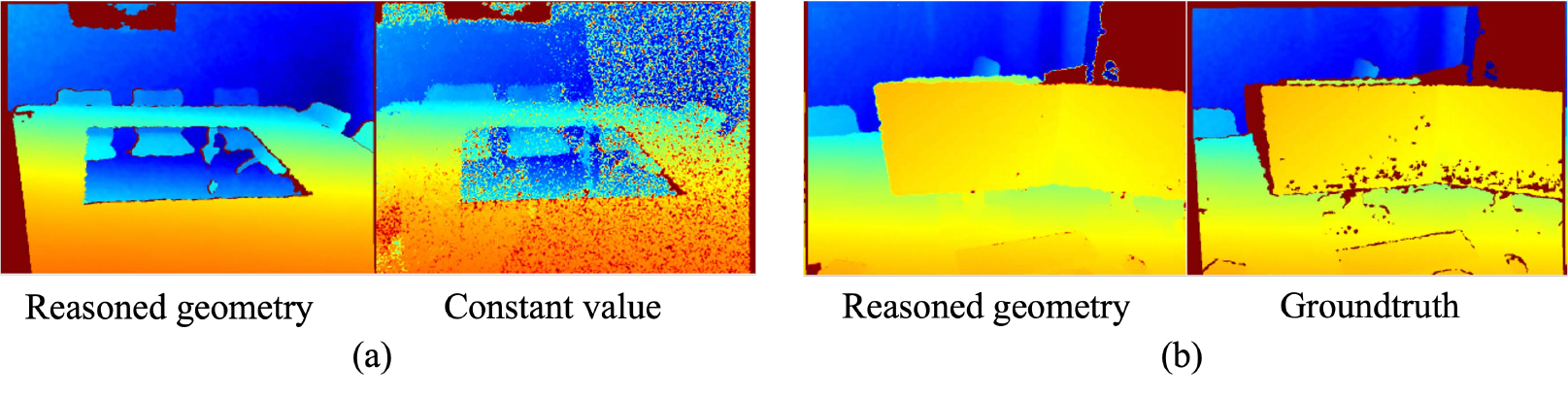}
    \vspace{-5mm}
    \caption{Comparison of the rendered depth map with or without reasoned geometric properties.}
    \label{fig: depthcompare}
    \vspace{-6mm}
\end{figure}

\noindent \textbf{Ablation of the geometric reasoning module}. To validate the significance of local geometric information, we remove the geometric reasoning module and assign some constant value to the properties of all Gaussian points. Specifically, we fix their opacity as 1, scale as (0.001, 0.001, 0.001), and rotation as (1, 0, 0, 0). Then we directly adopt these assigned properties to splat features. Fig.~\ref{fig: depthcompare}(a) gives an example of the comparison of the rendered depth with or without the geometric reasoning module. The depth map rendered by reasoned geometry has a smooth and accurate depth map generated by reasoned geometry, while the right depth map using a constant property-results in noisy and inaccurate geometry, especially around object boundaries.
This comparison highlights the importance of geometry reasoning for generating clean and reliable depth information.

Fig.~\ref{fig: depthcompare}(b) illustrates that even compared with the groundtruth captured by Lidar, which sometimes misses points on object surfaces, our method produces a more complete and cleaner depth map. This demonstrates the geometry reasoning ability of our approach.
Additionally, we evaluate the pretrained model of the two setups by the downstream embodied tasks, and the results are listed in Table~\ref{tabel: reasoned geometry}, which shows that incorporating reasoned geometry significantly improves task performance across both RLBench and ManiSkill2 benchmarks. Compared to the variant without geometry reasoning, our method achieves consistently higher success rates, with notable gains on challenging tasks like CloseJar.
\begin{table}[ht]
\centering
\caption{Ablation studies of the Geometric Reasoning Module.}
\label{tabel: reasoned geometry}
\small 
\setlength{\tabcolsep}{3pt} 
\resizebox{0.48\textwidth}{!}{
\begin{tabular}{lccccc}
 \hline
    & \multicolumn{3}{c}{RLBench} & \multicolumn{2}{c}{Maniskill} \\
    \cmidrule(lr){2-4} \cmidrule(lr){5-6}
    & TurnTap & CloseJar & MeatOffGrill & StackCube & Hang \\
 \hline
    No Reasoned Geo. & 0    & 0.24 & 0.72 & 0.28 & 0.92 \\
    Ours                 & \textbf{0.08} & \textbf{0.48} & \textbf{0.76} & \textbf{0.36} & \textbf{0.96} \\
 \hline
\end{tabular}}
\vspace{-3mm}
\end{table}

\noindent \textbf{Ablation of the Point-wise latent VAE module}. One of the most impactful components of our approach is the PL-VAE (point-wise latent VAE) applied in the PTv3 encoder’s latent space, enabling joint optimization of point-wise coordinates and features. We ablate this design by removing the PL-VAE and evaluating on all simulated datasets (RLBench and ManiSkill2); the results are included in Tables~\ref{tab:rlbench_recomputed} and~\ref{table: maniskill results}. As shown, eliminating the PL-VAE leads to substantial drops across all tasks. Although the model without PL-VAE still outperforms a PTv3-from-scratch baseline, it offers only a limited advantage over other 3D-aware methods. In contrast, reinstating the PL-VAE yields large, consistent gains and achieves the best overall performance, corroborating its role in learning compact, well-regularized structural latents.

\begin{figure}
    \centering
    \includegraphics[width=\linewidth]{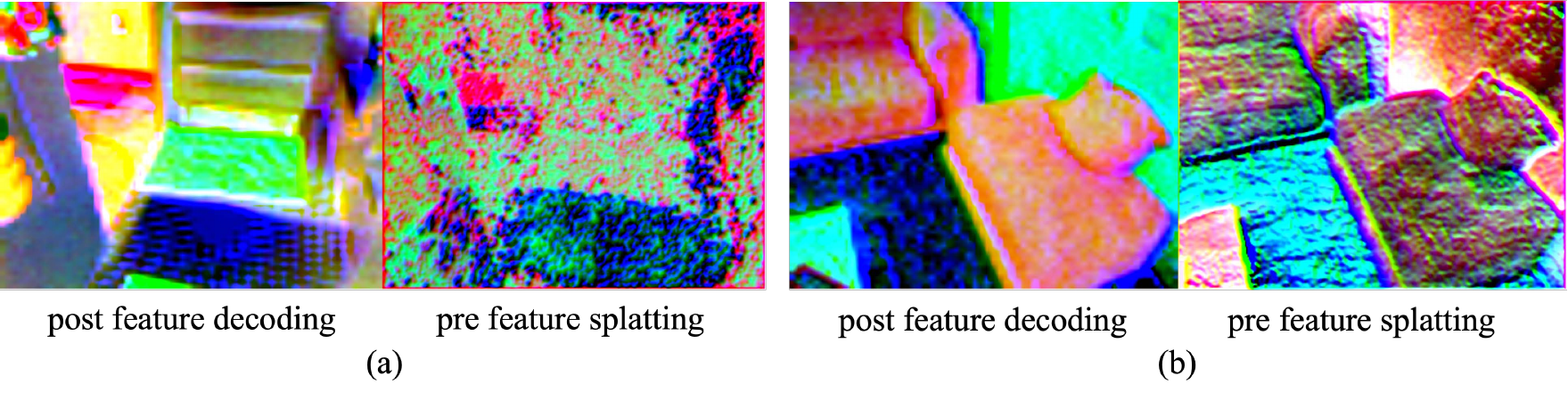}
    \vspace{-5mm}
    \caption{PCA visualizations of extracted features. The post feature decoding denotes the results of our feature-only rendering.}
    \label{fig: post feature decoding}
    \vspace{-3mm}
\end{figure}

\noindent \textbf{Ablation of the feature splatting}. As we discussed in Sec.~\ref{sec. main architecture}, typical methods splat both feature and color simultaneously, and build their respective loss terms, while we investigate and found that only feature terms provide the most contributions for representation learning. Here we provide a qualitative example in Fig.~\ref{fig: post feature decoding} to illustrate that. From this figure, we can observe that the features predicted by our solution exhibit clearer semantic separability, making distinct objects and boundaries more easily distinguishable. 
\noindent \textbf{Complexity Analysis}. We compare the training efficiency of our method with VC-1, SPA, and PonderV2, including the training GPU time and required device. Relative to prior 3D-aware baselines, our setup reduces aggregate GPU time by 4.6, 24, and 60 times, respectively. Just as importantly, it removes the need for large distributed clusters, simplifying reproduction and hyperparameter tuning by avoiding cross-GPU communication and schedule engineering. This smaller hardware footprint translates to lower operational cost and energy use. This advantage becomes even more pronounced when scaling up to larger datasets or training large models such as world models. Therefore, our method offers better scalability for large-scale pretraining.

\begin{table}[ht]
    \centering
    \renewcommand{\arraystretch}{1.2}
    \caption{Comparison of training setups for different methods.}
    \setlength{\tabcolsep}{6pt}
    \small
    \resizebox{0.48\textwidth}{!}{
    \begin{tabular}{lcccc}
        \hline
        & \textbf{VC-1} & \textbf{SPA} & \textbf{PonderV2} & \textbf{Ours} \\
         \hline
        \textbf{Device} 
            & - 
            & 80$\times$A100 
            & 8$\times$A100
            & 1$\times$A800 \\

        \textbf{Epoch} 
            & - 
            & 2000 
            & 100
            & 400 \\

        \textbf{GPU Time} 
            & $\sim$10000h 
            & $\sim$4000h 
            & $\sim$768h
            & $\sim$166h \\

         \hline
    \end{tabular}}
    \vspace{-5mm}
    \label{tab: training time}
\end{table}

\subsection{Evaluation on Real-world Experiments}
We deployed our pretrained representation encoder on a real-world robotic platform to further evaluate its generalization capabilities. Specifically, we utilized an Orbbec Femto Bolt time-of-flight camera to capture RGBD information, which is fed to the PTv3 encoder, Eq \ref{eq: plvae1}, and \ref{eq: regress vae gaussian parameters} to obtain an overall representation of the observation. The robotic system was equipped with an AgileX Piper robotic arm for policy execution. For the control policy, we employed ACT \cite{zhao2023learning}.
\begin{figure}
    \centering
    \includegraphics[width=0.49\textwidth]{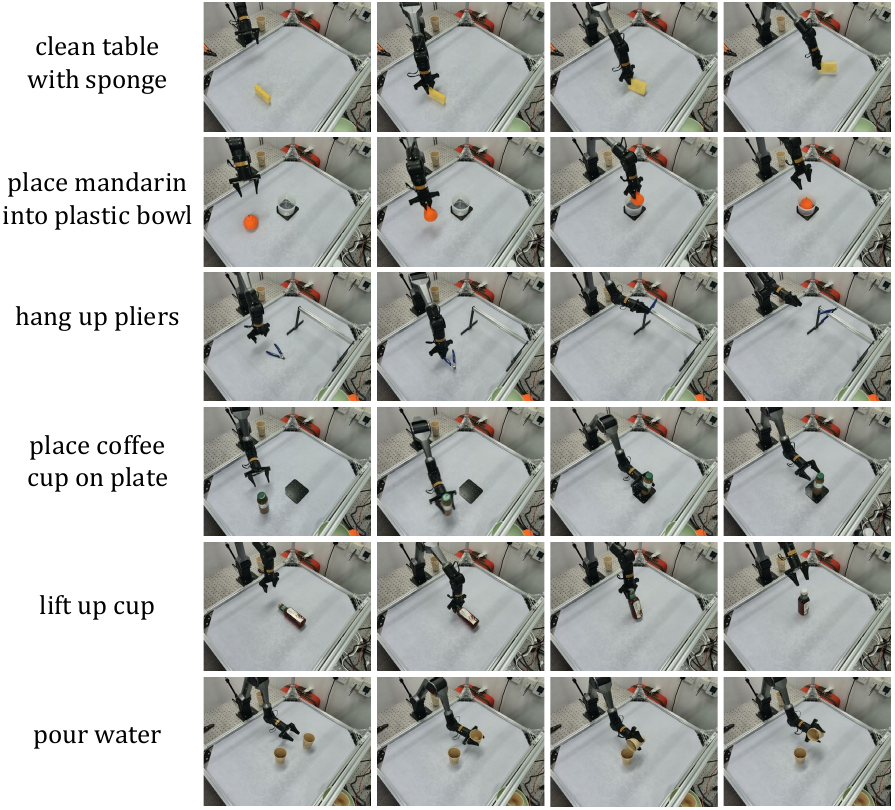}
    \caption{Real-world task demonstration of the proposed approach.}
    \label{fig: real_robot}
    \vspace{-3mm}
\end{figure}
Our experiments were conducted across six distinct manipulation tasks as shown in Fig. \ref{fig: real_robot}: (1) clean table with sponge, (2) place toy pea into plastic bowl, (3) hang up pliers, (4) place coffee cup on plate, (5) lift up cup, and (6) pour water. For each task, we collected expert demonstrations consisting of an action sequence spanning $360$ steps at $30$ Hz. The dataset was split into a training set of $40$ demonstrations and a test set of $10$ demonstrations. Training was performed on a single NVIDIA RTX3090 GPU for $10000$ epochs using the AdamW optimizer with a fixed learning rate of $5e-5$. The successful deployment of the trained model in real-world manipulation tasks underscores the practicality of our approach. Table~\ref{tab: real robot success rate} reports the success rate of some real robot tasks, where Ours (scratch) denotes our model without the pretraining. The results show that our model consistently outperforms the other conterparts. 
\begin{table}[t]
\centering
\caption{Success rate on three tasks across four methods}
\vspace{-3mm}
\begin{tabular}{lcccc}
\toprule
 & Pointnet & PonderV2 & Ours (scratch) & Ours \\
\midrule
place mandarin  & 56 & 72 & 76 & \textbf{84} \\
lift up bottle    & 64 & 48 & 40 & \textbf{68} \\
place starbucks & 36 & 32 & 28 & \textbf{36} \\
\bottomrule
\end{tabular}
\label{tab: real robot success rate}
\vspace{-3.5mm}
\end{table}
\vspace{-1mm}
\section{Discussion}
\noindent \textbf{Conclusion}. This paper introduces a 3D pretraining framework that fuses the strengths of implicit and explicit scene representations. By introducing a PL-VAE in the PTv3 latent space and regularizing both point features and coordinates, the method learns structural latent points that are compact, smooth, and geometry-aware. A streamlined 3DGS setup with predicted Gaussian properties, feature-only splatting, and depth alignment keeps supervision focused on representation quality, while an auxiliary point/color reconstruction stabilizes early training. Across RLBench and ManiSkill2, the approach achieves the best overall results among the compared 3D-aware methods and remains robust across tasks; ablations confirm that both the geometric reasoning module and the PL-VAE are key contributors. Finally, real-robot evaluations further support the method’s transferability beyond simulation.

\noindent \textbf{Limitation}. Since the method introduces a sampling step, it improves generalization ability but cannot represent object geometry with complete precision. Therefore, in scenarios that demand extremely high accuracy, such as surgical operations or threading a needle, its upper bound performance may be limited. Future research could explore 3D representations that couple deterministic and probabilistic components.
\vspace{-2mm}
\section*{acknowledgement}
\vspace{-1mm}
This work is partially supported by the National Natural Science Foundation of China (No. 62306261), HK RGC-Early Career Scheme (No. 24211525), ITSP Platform Project (No. ITS/600/24FP) and the SHIAE Grant (No. 8115074). This study was supported in part by the Centre for Perceptual and Interactive Intelligence, a CUHK-led InnoCentre under the Hong Kong SAR Government’s InnoHK initiative. This is also partially supported by Hong Kong RGC Strategic Topics Grant (No. STG1/E-403/24-N), and CUHK-CUHK(SZ)-GDST Joint Collaboration Fund (No. YSP26-4760949).

\bibliographystyle{IEEEtran}
\bibliography{References}
 

\end{document}